\documentclass[twocolumn, 11pt]{article}
\usepackage[a4paper,includeheadfoot,margin=2.54cm]{geometry}
\usepackage[font=footnotesize]{caption}
\usepackage[font=footnotesize]{subcaption}  % for displaying figures side by side

\usepackage{hyperref}  % urls, automatic in-document linking references
\hypersetup{  % to make hyperref package prettier
    colorlinks,
    linkcolor={red!50!black},
    citecolor={blue!50!black},
    urlcolor={blue!80!black}
}
\usepackage[table,xcdraw]{xcolor}  % background color in table
\usepackage{pgf}  % latex .pgf figures
\usepackage{tikz}  % plate diagrams
\usetikzlibrary{fit}
\usepackage{amsmath}  % more math commands, such as \begin{aligned}
\usepackage{amstext}  % for \text in mathmode
\usepackage{amssymb}  % for more math symbols, such as \mathbb
\DeclareRobustCommand{\rchi}{{\mathpalette\irchi\relax}}
\newcommand{\irchi}[2]{\raisebox{\depth}{$#1\chi$}} % inner command, used by \rchi
\usepackage{bm}
\newcommand{\figdir}{fig/}
\newcommand{\tikzdir}{\figdir tikz/}
\newcommand{\Dir}{\text{Dir}}
\newcommand{\Mult}{\text{Mult}}

\usepackage[T1]{fontenc}
\usepackage[utf8]{inputenc}
\usepackage{pdfpages}

\title{Viewpoint and Topic Modeling of Current Events}
\author{Kerry Zhang \\ kerryz@kth.se
    \and Jussi Karlgren \\ jussi@kth.se
    \and Cheng Zhang \\ chengz@kth.se
    \and Jens Lagergren \\ jensl@csc.kth.se}
\date{}

\begin{document}

% \hypersetup{pageanchor=false}
% \begin{titlepage}
% \clearpage\maketitle
% \thispagestyle{empty}
% \end{titlepage}

\maketitle

%\twocolumn
\begin{abstract}
\small
There are multiple sides to every story, and while statistical topic models have been highly successful at topically summarizing the stories in corpora of text documents, they do not explicitly address the issue of learning the different sides, the viewpoints, expressed in the documents. In this paper, we show how these viewpoints can be learned completely unsupervised and represented in a human interpretable form. We use a novel approach of applying CorrLDA2 for this purpose, which learns topic-viewpoint relations that can be used to form groups of topics, where each group represents a viewpoint. A corpus of documents about the Israeli-Palestinian conflict is then used to demonstrate how a Palestinian and an Israeli viewpoint can be learned. By leveraging the magnitudes and signs of the feature weights of a linear SVM, we introduce a principled method to evaluate associations between topics and viewpoints. With this, we demonstrate, both quantitatively and qualitatively, that the learned topic groups are contextually coherent, and form consistently correct topic-viewpoint associations.
\end{abstract}

\section{Introduction}
There is a wealth of text documents on the Web discussing current events, such as news articles, editorials, op-eds, etc. These events are commonly discussed from a subjective viewpoint, which is reflected in the word choices that are made by the authors. An example of this can be seen in these two sentences, written by two different authors about the Israeli-Palestinian conflict:

\begin{small}
%\itshape
\begin{enumerate}%[leftmargin=11pt]
    %\item The inadvertent killing by Israeli forces of Palestinian civilians – usually in the course of shooting at Palestinian terrorists – is considered no different at the moral and ethical level than the deliberate targeting of Israeli civilians by Palestinian suicide bombers.
    %\item In the first weeks of the Intifada, for example, Palestinian public protests and civilian demonstrations were answered brutally by Israel, which killed tens of unarmed protesters.
    %
    %\item ``harsh criticism of Yasir Arafat was to be expected, given PM Sharon's broad success in discrediting the Palestinian leader as a terrorist and a pathological liar.''
    %\item ``the rest of the world believes that this conflict is about the Israeli occupation and the struggle to end it, and that we are now seeing a rather typical process of decolonization.''
    %
    \item ``Those who hope that plans for separation would move us closer towards peace, or even calm the currently fierce struggle against occupation and the Israeli violence used to maintain that occupation, will be sorely mistaken.''
    \item ``Palestinians will be mistaken if they interpret this unilateral drive as a sign that terrorism is paying off, that Israelis are weak, and that yet more violence will produce more far-reaching gains.''
\end{enumerate}
\end{small}

A human with the necessary background knowledge can easily identify that sentence 1 is written from a Palestinian viewpoint and that sentence 2 is written from an Israeli viewpoint. We define viewpoint broadly as an attitude of mind, a standpoint that people take on various issues. The vocabulary is partitioned into different sets of words and we refer to each set as a modality. A topic is defined as a multinomial distribution over a modality of topical words, and an aspect is a multinomial distribution over a modality of opinion words. In this study, we show that topic-aspect relations can be learned and used to represent viewpoints.

Already in 1990, Deerwester et al. showed empirically that the co-occurrence structure of terms in text documents can be used to recover semantic meanings in latent topic structures \cite{Deerwester1990}. More recently, Blei et al. developed Latent Dirichlet Allocation (LDA) \cite{Blei2003a}, a statistical topic model that is now widely used for summarizing the topics that are discussed in a corpus of text documents. A more detailed description of LDA is provided in \autoref{subsec:model-lda}. While statistical topic models are able to infer the topics discussed in collections of text documents, they do not explicitly learn the viewpoints expressed in the documents. In this study, we show that topic-aspect relations can be learned and used to represent viewpoints. 

This ability to discover and represent viewpoints is of commercial interest. For instance, media monitoring and especially brand reputation management systems are sensitive to the formulation of targets and sentiments \cite{Karlgren2012}. The methodology developed in this study could provide a basis for more informed editorial oversight, and serve as a guidance for choosing which topics and viewpoints to monitor. A model which learns the dependence between viewpoints and topics could help avoid editorial bias in any subject where the public opinion is divided.

The contributions of this study are threefold.
\begin{enumerate}
    \item We improve the vocabulary partition used in previous related work to represent bimodal data of topical words and opinion words, resulting in more consistently correct topic-viewpoint associations.
    \item We apply CorrLDA2 \cite{Newman2006} to learn relations between topics and aspects. Using these relations, we form topics-aspect groups and show that each group represents a viewpoint.
    \item We introduce a principled method to evaluate which viewpoint a topic or aspect is associated with. This is done by leveraging the magnitudes and signs of the feature weights of a linear SVM, allowing us to quantitatively show that the inferred topic-aspect relations form consistently correct topic-viewpoint associations.
\end{enumerate}
\section{Related work}
Viewpoint modeling is a subfield of opinion mining and sentiment analysis, which aims to analyze opinionated documents to infer the sentiments expressed by people about various topics. It focuses on subjectivity and polarity, which in many applications may be more relevant to the end user than factual information.

Our goal is to learn both the topics and the viewpoints expressed in a corpus of text documents about current events, in a general and completely unsupervised setting. Hence, we focus on models that extend the statistical topic model LDA. No assumptions are made about specific document structures and we do not require any apriori knowledge of document-level viewpoint labels. An expressive model should be able to learn a large number of topics, while simultaneously being able to associate these topics with a lower number of viewpoints. Thus, a viewpoint is seen as spanning across multiple topics. Models proposed in previous studies have subsets of these properties, but none of the models satisfy all of them. In this section, we review the related work and highlight some of their capabilities and deficiencies.

% Consumer products
Several studies in extending LDA have been focused on extracting information about different aspects of consumer products \cite{Titov2008,Jo2011}. Although they are not directly related to the focus of modeling viewpoints in this paper, certain features employed in these models could be useful, such as having separate word-distributions for global topics and local aspects \cite{Titov2008}.

% Forums
Several studies focus on identifying viewpoints in user generated data with interdependent document structures \cite{Qiu2013a,Qiu2013}. However, these models depend on properties that are specific to online forums, such as threads, posts, users, and the interaction between users. Hence, these models are not applicable in a general setting. 

% ccLDA, CPT
Other studies use multiple collections of text documents as a basis for learning topics and viewpoints. Paul \& Girju let each collection of documents represent a culture, and introduce cross-collection LDA (ccLDA) \cite{Paul2009} to learn the similarities and differences in topics between cultures. Fang et al. introduce the Cross-Perspective Topic Model (CPTM) \cite{Fang2012}, which jointly learns topics and collection-specific opinions. Two modalities of data consisting of topical- and opinion words are introduced by separating the vocabulary based on part-of-speech: nouns represent topic words, and opinion words are represented by adjectives, verbs, and adverbs. In both these models, the viewpoint of a document is assumed to be known a priori since the viewpoints are determined by which text collection a document belongs to, but our goal is to learn these viewpoints in an unsupervised setting.

% TAM, JTV, VODUM
An unsupervised approach is introduced by Paul \& Girju through the Topic-Aspect Model (TAM) \cite{Paul2010a}, which jointly discovers topics and aspects that represent viewpoints . However, in the generative process of this model, the topic mixture and aspect mixture of a document are sampled independently of each other. As such, it does not model a relationship between topics and aspects. In fact, topics and aspects are viewed as two orthogonal dimensions.

The relations between topics and viewpoints are learned in the Joint Topic Viewpoint model (JTV) \cite{Trabelsi2014}. However, no relation between viewpoints across topics is learned, resulting in $K \cdot L$ different topic-viewpoint pairs, where $K$ is the number of topics and $L$ is the number of viewpoints. The authors suggest applying constrained k-means clustering algorithm as a post-processing step to find $L$ clusters of viewpoints. We aim to introduce viewpoints that are explicitly modeled as spanning across topics.

The Viewpoint and Opinions Discovery Unification Model (VODUM) \cite{Thonet2016} uses the same approach of creating a bimodal dataset based on part-of-speech as Fang et al. \cite{Fang2012}, resulting in a set of topical words, and set of opinion words.~Unlike the previously mentioned models, including LDA, this model does not learn a document-specific topic distribution, which could limit its ability to learn topics on corpora of multi-subject documents. Additionally, it models $T \cdot V$ different opinion word-distributions, where $V$ is the number of viewpoints. Again, our aim is to model only $V$ viewpoints that span across topics. % This implicitly assumes that the corpus under investigation only consists of documents that are about a single subject, such as the Israeli-Palestinian conflict.%In this paper, we want to introduce a model that is able to generalize to any corpus consisting of any number of subjects. % This implicitly assumes that the corpus under investigation only consists of documents that are about a single subject, such as the Israeli-Palestinian conflict.%In this paper, we want to introduce a model that is able to generalize to any corpus consisting of any number of subjects.

% corrLDA2
% Extensions to LDA have also been developed outside the field of viewpoint modeling. For the task of entity-topic modeling, Newman et al. introduce CorrLDA2 \cite{Newman2006}, which is based on Corr-LDA \cite{Blei2003b} by Blei \& Jordan, to model the relationship between the two modalities of named entities (persons, organizations, locations) and the rest of the vocabulary. Other studies use CorrLDA2 on the modalities of emotional words (e.g. ``love'', ``hate'') and the rest of the vocabulary \cite{Carlson2013}. Using the relations learned by CorrLDA2 between the latter modalities, groups of topics are formed based on the emotions associated with them. Hence, CorrLDA2 is able model a large number of topics and their associations with a low number of basic emotions. These basic emotions are seen as spanning across multiple topics, and they are learned completely unsupervised. Using this approach, if viewpoints can be learned instead of basic emotions, then CorrLDA2 would have all the required properties mentioned in the beginning of this section, making it a good candidate for the task of viewpoint and topic modeling.

Based on Corr-LDA \cite{Blei2003b}, CorrLDA2 \cite{Newman2006} is originally introduced for the unrelated task of entity-topic modeling, which addresses the textual interactions between who/where, i.e. named entities (persons, organizations, locations) and what, i.e. the topics. This is achieved by partitioning the vocabulary into one modality consisting of named entities, and another modality consisting of the rest of the vocabulary. Separate multinomial distributions are then learned over the two modalities, and inter-modality relations are also learned. As an example, the model infers that topics related to the September 11 attacks are related to the the entities of the World Trade Center, New York City, etc.

Carlson et al. \cite{Carlson2013} use the relations that CorrLDA2 learns to group topics based on the basic emotions that the topics are associated with. This is achieved by partitioning the vocabulary into the modalities of emotional words (e.g. ``love'', ``hate'') and the rest of the vocabulary. For instance, the topics of politics, economy, and terrorism are found to be associated with the basic emotion of hate, while topics such as family and relationships are associated with love. Hence, CorrLDA2 is able to model a large number of topics, and their associations with a low number of basic emotions. These basic emotions are seen as spanning across multiple topics, and they are learned completely unsupervised. Using this approach, if viewpoints can be learned instead of basic emotions, then CorrLDA2 would be a good candidate for the task of viewpoint and topic modeling.

% Non-LDA approaches
Other approaches to viewpoint modeling that are not based on LDA also exist. Lin et al. build upon the Naive Bayes model \cite{Lin2006}. Park et. al introduce a graph-based approach based on a modified version of the HITS algorithm and perform viewpoint classification using SVM \cite{Park2011}.
\section{Model}
In short, Newman et al. \cite{Newman2006} introduce CorrLDA2 to model the \emph{relations} between entities and topics. Carlson et al. \cite{Carlson2013} use the learned relations to form groups of topics based on their associations with basic \emph{emotions}. In this paper, we show that CorrLDA2 can be used to group topics based on which \emph{viewpoint} they are associated with.

In contrast to previous work on viewpoint and topic modeling, our approach does not assume the documents' viewpoints to be known a priori, relationships between topics and aspects are learned, and aspects can be associated with multiple topics. Unlike Corr-LDA, CorrLDA2 does not require the number of multinomial distributions over the two separate modalities to be equal, allowing us to model relationships between a large number of topics with a lower number of aspects.

Both LDA and CorrLDA2 are statistical latent variable graphical models. In these models, a topic is modeled as a multinomial distribution over a vocabulary of words, where words that are related to a particular topic are assigned higher probabilities than words that are not related to the topic. For instance, one topic might assign high probabilities to the words ``occupation'', ``violence'', ``illegal'', ``right'', etc. which then could be interpreted as a topic about the Israeli occupation of what the Palestinian consider to be their territory. A document is viewed as a mixture of these topics and each word in the document is considered to be assigned to one of these topics. The words in the documents are observed variables and the topic assignment of each word is an unobserved, latent variable. 

In LDA and CorrLDA2, all of the distributions, whether it is the per-topic word-distributions, per-document topic-distribution, or in the case of CorrLDA2, the per-topic aspect-distribution, introduced in \autoref{subsec:model-corrlda2}, are modeled by the multinomial distribution. Since these are Bayesian models, a Dirichlet distribution, a conjugate prior to the multinomial distribution, is introduced for each of the aforementioned multinomial distributions.
% We will investigate how the relations between two modalities of data learned by CorrLDA2 can be used to model the relationship between topics and viewpoints.

\subsection{Modalities: Vocabulary Partition}
In previous studies \cite{Fang2012,Thonet2016}, a bimodal data set is created from a corpus of text documents by partitioning the vocabulary into a set of topical words and a set of opinion words. They base this partition on part-of-speech categories, with nouns representing the modality of topical words, while adjectives, verbs, and adverbs represent the modality of opinion words.  We apply CorrLDA2 to these two modalities as well, with the addition of including named entities to the set of opinion words. This is motivated by the observation made in \autoref{subsec:results-corrlda2} that people tend to talk about themselves, and we also quantitatively show that this leads to more consistently correct topic-viewpoint associations in \autoref{subsec:results-consistency}. We refer to this partition as (opinion+ne), and other partitions are also investigated, such as using adjectives and named entities (adj+ne) as opinion words with the rest of the vocabulary as topical words.

\begin{table*}[ht!]
\renewcommand{\arraystretch}{1.3}
\centering
    \caption{Notation used for LDA and CorrLDA2.}
    \label{tab:notation}
    \centerline{
    \begin{tabular}{|c | l|}
    \hline
    $W, \tilde{W}$                      & Size of the vocabulary for the set of topical words, and the set of opinion words \\
    $T$, $\tilde{T}$                    & Number of topics, and number of aspects \\
    $D$                                 & Number of documents in the corpus \\
    $N_{w, d}, N_{\tilde{w}, d}, N_d$   & Number of topical words, and number of opinion words in document $d$. $N_d = N_{w, d} + N_{\tilde{w}, d}$ \\ 
    \hline
    $w_i, \tilde{w}_{\tilde{i}}, v_j$   & Topical word, opinion word, and any word (used in LDA) \\
    $z_i, \tilde{z}_{\tilde{i}}$        & Topic assigned to topical word $w_i$, and aspect assigned to opinion word $\tilde{w}_{\tilde{i}}$ \\
    $x_{\tilde{i}}$                     & $\in \{1, \ldots, T\}$, supertopic assigned to opinion word $\tilde{w}_{\tilde{i}}$ \\
    \hline
    $\phi_t, \tilde{\phi}_{\tilde{t}}$  & Topical word-distribution of topic $t$, and opinion word-distribution of aspect $\tilde{t}$ \\
    $\psi_t$                            & Aspect-distribution of topic $t$ \\
    $\theta_d$                          & Topic-distribution of document $d$ \\
    \hline
    $\alpha, \beta, \tilde{\beta}, \gamma$ & Symmetric Dirichlet hyperparameters \\
    \hline
    \end{tabular}
    }
\end{table*}

\subsection{LDA}
\label{subsec:model-lda}

\begin{figure}[ht]
\vspace{-5pt}
\centering
\scalebox{0.75}{\pgfdeclarelayer{background}
\pgfdeclarelayer{foreground}
\pgfsetlayers{background,main,foreground}

\begin{tikzpicture}

\tikzstyle{surround} = [thick,draw=black,rounded corners=1mm]

% Define styles
\tikzstyle{scalarnode} = [circle, draw, fill=white!11,  
    text width=1.2em, text badly centered, inner sep=2.5pt]

\tikzstyle{scalarnodeCyan} = [circle, draw=cyan, fill=white!11,  
    text width=1.2em, text badly centered, inner sep=2.5pt]
\tikzstyle{discnode}=[rectangle,draw,fill=white!11,minimum size=0.9cm]

\tikzstyle{Vnode}=[circle, radius=1pt,draw,fill=black]
\tikzstyle{vectornode} = [circle, draw, fill=white!11,  
    text width=2.3em, text badly centered, inner sep=2pt]
\tikzstyle{state} = [rectangle, draw, text centered, fill=white, 
    text width=8em, text height=6.7em, rounded corners]

\tikzstyle{arrowline} = [draw,color=black, -latex]
\tikzstyle{carrowline} = [line width=2pt, draw,color=black, -latex]
\tikzstyle{line} = [draw]

% Nodes
\node [Vnode] at ( 0.5, 1.5) (alpha_s){};
\node [] at ( 0, 1.5) (){$\alpha$};
\node [scalarnode] at ( 2, 1.5 ) (theta) { $\theta$ };
\node [scalarnode] at ( 4, 1.5) (z) {$z$};
\node [scalarnode, fill=black!30] at ( 5.5, 1.5) (w) {$v$};
\node [scalarnode] at ( 8, 1.5) (phi) {$\phi$};
\node [Vnode] at ( 9.5, 1.5) (beta) {};
\node [] at ( 10, 1.5) () {$\beta$};
% aspects

%Plates
\node[surround, inner sep = .33cm] (f_N) [fit = (z)(w) ] {};
\node[surround, inner sep = .4cm] (f_M) [fit = (f_N)(theta)] {};
\node[surround, inner sep = .33cm] (f_phi) [fit = (phi)] {};

% Annotate each plate
\node [] at (6.4, 0.60) (D) {\scriptsize $D$};
\node [] at (5.9, 1) (N) {\scriptsize $N_d$};
\node [] at (8.55, .95) () {\scriptsize $T$};

% Connections 
\path [arrowline] (alpha_s) to (theta); 
\path [arrowline] (theta) to (z); 
\path [arrowline] (z) to (w); 
\path [arrowline] (phi) to (w); 
\path [arrowline] (beta) to (phi); 

\end{tikzpicture}}
%\vspace{20pt}
\caption{Graphical representation of LDA}
\vspace{-3pt}
\label{fig:ldaPGM}
\end{figure}
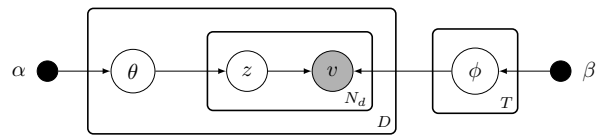

The graphical model of LDA is shown in \autoref{fig:ldaPGM} and the notation is explained in \autoref{tab:notation}. LDA is a generative statistical model and it models the creation of a document through the following generative process:

\begin{enumerate}
    \item For all $t = 1, \ldots, T$ topics, sample a topic-word distribution $\phi_t \sim \Dir(\beta)$
    \item For all $d = 1, \ldots, D$ documents, sample a document-topic distribution $\theta_d \sim \Dir(\alpha)$
    \item For each of the $N_d$ words $v_j$ in document $d$:
    \begin{enumerate}
        \item Sample a topic $z_j \sim \Mult(\theta_d)$
        \item Sample a word $v_j \sim \Mult(\phi_{z_j})$
    \end{enumerate}
\end{enumerate}

Note that LDA does not distinguish between topical words and opinion words. By separating these two modalities, CorrLDA2 is able to model both topics and aspects, as well as the relations between them.

\subsection{CorrLDA2}
\label{subsec:model-corrlda2}

\begin{figure}[ht]
\vspace{-5pt}
\centering
\scalebox{0.8}{\pgfdeclarelayer{background}
\pgfdeclarelayer{foreground}
\pgfsetlayers{background,main,foreground}

\begin{tikzpicture}

\tikzstyle{surround} = [thick,draw=black,rounded corners=1mm]

% Define styles
\tikzstyle{scalarnode} = [circle, draw, fill=white!11,  
    text width=1.2em, text badly centered, inner sep=2.5pt]

\tikzstyle{scalarnodeCyan} = [circle, draw=cyan, fill=white!11,  
    text width=1.2em, text badly centered, inner sep=2.5pt]
\tikzstyle{discnode}=[rectangle,draw,fill=white!11,minimum size=0.9cm]

\tikzstyle{Vnode}=[circle, radius=1pt,draw,fill=black]
\tikzstyle{vectornode} = [circle, draw, fill=white!11,  
    text width=2.3em, text badly centered, inner sep=2pt]
\tikzstyle{state} = [rectangle, draw, text centered, fill=white, 
    text width=8em, text height=6.7em, rounded corners]

\tikzstyle{arrowline} = [draw,color=black, -latex]
\tikzstyle{carrowline} = [line width=2pt, draw,color=black, -latex]
\tikzstyle{line} = [draw]

% Nodes
\node [Vnode] at ( 0.6, 5.2) (alpha_s){};
\node [] at ( 0, 5.2) (){$\alpha$};
\node [scalarnode] at ( 2.5, 5.2 ) (theta) { $\theta$ };
\node [scalarnode] at ( 4.3, 5.2) (z) {$z$};
\node [scalarnode, fill=black!30] at ( 5.7, 5.2) (w) {$w$};
\node [scalarnode] at ( 8, 5.2) (phi) {$\phi$};
\node [Vnode] at ( 9.5, 5.2) (beta) {};
\node [] at ( 10, 5.2) () {$\beta$};
% aspects
\node [] at ( 2.3, 1.5) (){$\gamma$};
\node [Vnode] at ( 2.8, 1.5) (gamma){};
\node [scalarnode] at ( 4.3, 1.5) (psi) {$\psi$};
\node [scalarnode] at ( 3.0, 3.6) (x) {$x$};  % 3.6 -> 3.6
\node [scalarnode] at ( 4.3, 3.6) (z_a) {$\tilde{z}$};
\node [scalarnode, fill=black!30] at ( 5.7, 3.6) (w_a) {$\tilde{w}$};
\node [scalarnode] at ( 8, 3.6) (phi_a) {$\tilde{\phi}$};
\node [Vnode] at ( 9.5, 3.6) (beta_a) {};
\node [] at ( 10, 3.6) () {$\tilde{\beta}$};

%Plates
\node[surround, inner sep = .33cm] (f_N_t) [fit = (z)(w) ] {};
\node[surround, inner sep = .33cm] (f_N_a) [fit = (x)(z_a)(w_a) ] {};
\node[surround, inner sep = .32cm] (f_phi) [fit = (phi)] {};
\node[surround, inner sep = .32cm] (f_phi_a) [fit = (phi_a)] {};
% CorrLDA2
\node[surround, inner sep = .32cm] (f_phi) [fit = (psi)] {};
\node[surround, inner sep = .33cm] (f_M) [fit = (theta)(f_N_t)(f_N_a)] {};

% Annotate each plate
\node [] at (4.84, 0.96) (T_A) {\scriptsize $T$};
\node [] at (6.55, 2.70) (D) {\scriptsize $D$};
\node [] at (8.55, 4.65) (T) {\scriptsize $T$};
\node [inner sep = 0.5pt] at (3.850, 4.750) (N_t) {\scriptsize $N$};
\node [] at (4.14, 4.68) (N_t_sub) {\scriptsize $_{w, d}$};
% CorrLDA2
\node [] at (8.55, 3.06) (A) {\scriptsize $\tilde{T}$};
\node [] at (2.70, 3.1) (N_a) {\scriptsize $N_{\tilde{w}, d}$};

% Connections 
\path [arrowline] (alpha_s) to (theta);
\path [arrowline] (theta) to (z); 
\path [arrowline] (z) to (w); 
\path [arrowline] (phi) to (w); 
\path [arrowline] (beta) to (phi);
% CorrLDA2
\path [arrowline] (gamma) to (psi);
\path [arrowline] (psi) to (z_a);
\path [arrowline] (N_t) to (x);
\path [arrowline] (x) to (z_a); 
\path [arrowline] (z_a) to (w_a); 
\path [arrowline] (phi_a) to (w_a); 
\path [arrowline] (beta_a) to (phi_a);

\end{tikzpicture}}
%\vspace{20pt}
\caption{Graphical representation of CorrLDA2}
\vspace{-3pt}
\label{fig:corrlda2PGM}
\end{figure}
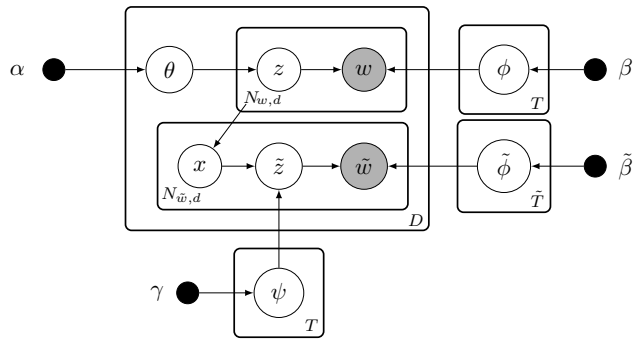

The topics inferred by CorrLDA2 are learned in the same way that LDA learns topics, with the difference that the multinomial probability distributions of the topics are only over the set of topical words. Separate multinomial probability distributions are then formed over the set of opinion words, which we refer to as aspects. CorrLDA2 also models relations between topics and aspects by introducing a multinomial per-topic aspect distribution $\psi_t$. In the generative process, topics are first sampled for the topical words in a document. Then, for each opinion word $\tilde{w}_{\tilde{i}}$, a supertopic $x_{\tilde{i}} \in \{1,\ldots,T\}$ is sampled based on only the topics associated with the topical words in the document. Conditioned on this supertopic, an aspect $\tilde{z}_{\tilde{i}}$ is sampled from the topic-aspect distribution $\psi_{x_{\tilde{i}}}$. By sampling the supertopic only from the topics associated with the topical words in the same document, a correspondence between topics and aspects is enforced. The introduction of a topic-aspect distribution $\psi_t$ also allows for different number of topics $T$ and aspects $\tilde{T}$.

The intuition is that viewpoints are often associated with groups of entities, as well as opinions that are expressed through characteristic word choices. These entities and opinionated word choices are used consistently, spanning across topics, and we model them as aspects. Furthermore, topics tend to be be associated with viewpoints. For instance, Palestinians are more likely to bring up the topic of resistance movements against what they describe as illegal occupation, while Israelis are more likely to bring up the topic of security from what they describe as violent terrorist attacks, as shown in \autoref{subsec:results-topic-aspect-relations}. Aspects and topics that are associated with the same viewpoint thus tend to co-occur, which is modeled by the topic-aspect distribution $\psi_t$. Hence, in our approach, a viewpoint is represented by both an aspect as well as the topics related to the aspect.

The graphical model of CorrLDA2 is shown in \autoref{fig:corrlda2PGM}, and its generative process is as follows:

\begin{flushleft}
\begin{enumerate}
    \item For all $t = 1, \ldots, T$ topics, sample a topic-topical word distribution $\phi_t~\sim~\Dir(\beta)$, and a topic-aspect distribution $\psi_t~\sim\Dir(\gamma)$
    \item For all $\tilde{t} = 1, \ldots, \tilde{T}$ aspects, sample an aspect-opinion word distribution $\tilde{\phi}_{\tilde{t}}~\sim~\Dir(\tilde{\beta})$
    \item For all $d = 1, \ldots, D$ documents, sample a document-topic distribution $\theta_d~\sim~\Dir(\alpha)$
    \item For each of the $N_{w, d}$ topical words $w_i$ in document $d$:
    \begin{enumerate}
        \item Sample a topic $z_i \sim \Mult(\theta_d)$
        \item Sample a topical word $w_i \sim \Mult(\phi_{z_i})$
    \end{enumerate}
    \item For each of the $N_{\tilde{w}, d}$ opinion words $\tilde{w}_{\tilde{i}}$ in document $d$:
    \begin{enumerate}
        \item Sample a supertopic $x_{\tilde{i}} \sim \text{Unif}(z_{w_1}, \ldots, z_{w_{N_{w, d}}})$
        \item Sample an aspect $\tilde{z}_{\tilde{i}} \sim \Mult(\psi_{x_{\tilde{i}}})$
        \item Sample an opinion word $\tilde{w}_{\tilde{i}} \sim \Mult(\tilde{\phi}_{\tilde{z}_{\tilde{i}}})$
    \end{enumerate}
\end{enumerate}
\end{flushleft}

A linear SVM is used for evaluating which viewpoint the topics and aspects are associated with by interpreting the corresponding feature weight, as described in \autoref{subsubsec:svm-feature-weights}. However, we want to emphasize that the SVM is only used for evaluation and is \emph{not} part of the learning process of forming representations of viewpoints, which is completely unsupervised. In CorrLDA2, these viewpoint representations are learned by forming groups of topics with an aspect, based on the inferred topic-aspect relations.

\subsection{Inference and Parameter Estimation}
In LDA, the target of inference is $p(\mathbf{z} \mid \mathbf{w})$, i.e. the joint probability of all topic assignments, $\mathbf{z}$, conditioned on the observed words, $\mathbf{w}$, in all of the documents of the corpus. Exact inference of this high-dimensional posterior distribution is intractable. Hence, Gibbs sampling, a Markov chain Monte Carlo method, is used for approximate inference. In Gibbs sampling, instead of directly sampling from the complex joint distribution of all of the topic assignments, only one of topic assignments, $z_i \in \mathbf{z}$, is sampled at a time, while holding all the other parameters constant. This is achieved using the full conditional distribution $p(z_i \mid \mathbf{z}_{-i}, \mathbf{w})$, where $\mathbf{z}_{-i}$ refers to all topic assignments $\mathbf{z}$ except $z_i$. Additionally, a collapsed Gibbs sampler can be obtained by integrating out the variables $\theta$ and $\phi$. For a full description of the process of Bayesian inference of LDA, we refer the reader to a study due to Heinrich \cite{Heinrich2008}.

Approximate inference using collapsed Gibbs sampling is also used for CorrLDA2. Using the same notation introduced by Newman et al. \cite{Newman2006}, we iteratively sample from the following full conditional distributions:

\begin{equation}
\begin{aligned}
    p(z_i = t &\mid w_i = w, \mathbf{z}_{-i}, \mathbf{w}_{-i}, \alpha, \beta) \propto \\
              &\frac{C^{TD}_{td, -i} + \alpha}{\sum_{t'} C^{TD}_{t'd, -i} + T\alpha} \;
               \frac{C^{WT}_{wt, -i} + \beta}{\sum_{w'} C^{WT}_{w't, -i} + W\beta}
\end{aligned}
\end{equation}

\noindent where $C^{TD}_{td, -i}$ is the number of words assigned to topic $t$ in document $d$, excluding word $i$. $C^{WT}_{wt, -i}$ is the number of times the topical word $w$ is assigned to topic $t$, excluding word $i$. For the joint, full conditional probability of a supertopic and aspect, we have:

\begin{equation}
\begin{aligned}
    p&(\tilde{z}_{\tilde{i}} = \tilde{z}, x_{\tilde{i}} = t \mid \tilde{w}_{\tilde{i}} = \tilde{w}, \tilde{\mathbf{z}}_{-\tilde{i}}, \mathbf{z}, \tilde{\mathbf{w}}_{-\tilde{i}}, \alpha, \tilde{\beta}) \propto \\
        &\frac{C^{TD}_{td}}{N_{w, d}} \;
         \frac{C^{\tilde{T}T}_{\tilde{t}t, -{\tilde{i}}} + \gamma}{\sum_{\tilde{t}'} C^{\tilde{T}T}_{\tilde{t}'d, -{\tilde{i}}} + \tilde{T\gamma}} \;
         \frac{C^{\tilde{W}\tilde{T}}_{\tilde{w}\tilde{t}, -{\tilde{i}}} + \tilde{\beta}}{\sum_{\tilde{w}'} C^{\tilde{W}\tilde{T}}_{\tilde{w}'\tilde{t}, -{\tilde{i}}} + \tilde{W}\tilde{\beta}}
\end{aligned}
\end{equation}

\noindent where $C^{\tilde{T}T}_{\tilde{t}t, -{\tilde{i}}}$ is the number of times aspect $\tilde{t}$ and topic $t$ are assigned to the same opinion word, excluding word $\tilde{i}$. $C^{\tilde{W}\tilde{T}}_{\tilde{w}\tilde{t}, -{\tilde{i}}}$ is the number of times the opinion word $\tilde{w}$ is assigned to aspect $\tilde{t}$, excluding word $\tilde{i}$.
\section{Methodology}

\subsection{Dataset}
The dataset used in our experiments consists of articles published on the Bitterlemons website\footnote{http://www.bitterlemons.net/}. The stated purpose of the website is to ``reflect a joint Palestinian-Israeli effort to promote a civilized exchange of views about the Israel-Arab conflict and additional Middle East issues among a broad spectrum of participants.''\footnote{http://www.bitterlemons.net/about.php} The articles in the corpus were published during the years 2001-2005 and were collected by Wei-Hao Lin et al. \cite{Lin2006}. Every week during this period, one event or topic related to the Israel-Arab (predominantly Palestinian) conflict was discussed, and an Israeli and a Palestinian editor each contributed an article to the discussion. In addition to these same two editors, one Israeli and one Palestinian guest were also invited to contribute to the discussion. Thus, weekly editions of four articles were published, and 594 articles were collected in total. 

There are two main reasons for choosing this dataset. First, viewpoint labels on a document level can be extracted since the author's nationality is known. Second, it reflects a realistic setting where the documents are written by a small number of authors (two editors), as well as a larger group of authors (more than 200 different guest authors).

Pre-processing of the documents is required to extract part-of-speech (POS) tags. The Stanford POS Tagger\footnote{http://nlp.stanford.edu/software/tagger.shtml} is used for this purpose and only tokens in the categories noun, adjective, adverb, and verb are kept. The Stanford Named Entity Recognizer\footnote{http://nlp.stanford.edu/software/CRF-NER.shtml} is used to extract named entities. To reduce word sparsity, all tokens are downcased, and the WordNet Lemmatizer\footnote{http://www.nltk.org/\_modules/nltk/stem/wordnet.html} is also used to lemmatize the tokens. We found that using a lemmatizer instead of a stemmer, which cannot discriminate between words which have different meanings depending on part-of-speech, increases the classification accuracy for all models.

\subsection{Quantitative evaluation}
\label{sec:quantitative-evaluation}
\subsubsection{Document Classification Accuracy}
LDA assigns one of $T$ number of topics to every word in a document. Hence, a document can be represented by a $T$-dimensional vector, where the $t^\text{th}$ dimension (also called a feature) of the vector represents the fraction of words in the document assigned to topic $t$. In CorrLDA2, $T$ topics and $\tilde{T}$ aspects are used. Thus, each document can be represented by a $(T + \tilde{T})$-dimensional vector. Using these vector-representations of each document instead of e.g. a bag-of-words representation of a document, which has a dimensionality equal to the size of the vocabulary, results in a huge dimensionality reduction. To quantitatively evaluate whether these topics or aspects are informative of the viewpoint expressed in a document, an SVM can be trained using the $T$-dimensional or the $(T + \tilde{T})$-dimensional vector representation of the documents, using the nationality of each document's author as the label. Using this supervised classification method, combined with 5-fold cross-validation to evaluate the models' performances on unseen data, an average document classification accuracy can be obtained. The more informative the topics or aspects are of an author's viewpoint is, the higher document classification accuracy can be achieved. The dataset is completely balanced, i.e. there are an equal amount of Israeli as there are Palestinian documents. Hence, the random guess accuracy is $0.5$, and accuracy, defined as the number of correctly classified documents divided by the total number of documents, is thus a sufficient evaluation criteria. 

% Also, since only non-negative feature weights are used (each feature is a fraction, having a value between 0 and 1), the sign of the feature weight also signifies whether the feature contributes to a negative (Palestinian, which we label as $-1$) classification or a positive (Israeli, label $1$) classification. This can be seen in the in the formulation of the hard-margin primal optimization problem
\subsubsection{SVM Feature Weights}
\label{subsubsec:svm-feature-weights}
Since a linear SVM is used and all feature values are fractions, i.e. normalized with values between 0 and 1, the magnitude of a feature weight in the primal optimization problem reflects how informative the feature is for the classification process \cite{chang2008,guyon2002}. Also, since we perform binary classification with only non-negative feature values, the sign of the feature weight indicates whether a feature contributes to a Palestinian or an Israeli classification. This can be seen in the formulation of the hard-margin primal optimization problem

\begin{equation}
\begin{aligned}\label{eq:svm}
& {\text{minimize}}
& & \frac{1}{2}\mathbf{w}^T \mathbf{w} \\
& \text{subject to}
& & y_d (\mathbf{w}^T \bm{\rchi}_d - b) \geq 1, \quad d = 1, \ldots, D
\end{aligned}
\end{equation}

\noindent where $y_d \in \{-1, +1\}$ is the label of document $d$, with $-1$ being the Palestinian label and $+1$ being the Israeli label. $\bm{\rchi}_d$ is the feature representation of document $d$, with $\bm{\rchi} \in \mathbb{R}^T$ for LDA and $\bm{\rchi} \in \mathbb{R}^{(T+\tilde{T})}$ for CorrLDA2. $\mathbf{w}$ is the vector of feature weights, and $b$ determines the offset of the maximum-margin hyperplane. In the constraint, we observe that when all feature values $\rchi \in \bm{\rchi}$ are non-negative (which they are in our case, since each feature is a fraction), negative feature weights $w \in \mathbf{w}$ contribute to a classification of $y = -1$ (Palestinian), while positive feature weights contribute to a classification of $y=+1$ (Israeli). A similar analysis can be made for the soft-margin case, where a loss function, usually L1- or L2-loss, is included in the optimization problem of \autoref{eq:svm}. Chang \& Lin show that in general, the choice between L1- and L2-loss does not affect the process of feature selection, where feature importance is determined by the magnitude of the feature weight learned by a linear SVM \cite{chang2008}. For our experiments, we use the L2-loss function with a linear SVM from the Python library scikit-learn \cite{scikit-learn}.

\paragraph{Perplexity} Another common method used for evaluating how well a topic model is able to learn from the dataset is its perplexity obtained on unseen data. The perplexity is a weighted geometric average of the inverses of the word-probabilities. However, since CorrLDA2 partitions the vocabulary into two separate vocabularies, the word-probabilities cannot be compared with models that do not partition the vocabulary, such as LDA. Hence, perplexity is not used to evaluate the models. 

\subsection{Qualitative evaluation}
Another property that is of practical importance is how coherent and informative the learned topics or aspects seem to a human observer. This is by definition a subjective evaluation and is conducted in conjunction with the quantitative evaluation methods to incorporate a sense of objectivity. We investigate how the topic-aspect relations learned by CorrLDA2 can assist a human expert in identifying viewpoints expressed in the corpus.

\subsection{Parameter Settings}
For the experiments, the Dirichlet hyperparameters are set as follows: $\alpha=0.1$, $\beta=0.01$, $\tilde{\beta}=0.01$, $\gamma=0.01$. The reason for a low value of $\gamma$ is to encode a sparse topic-aspect distribution. The number of iterations used in the Gibbs sampling process of LDA is set to 600, and 2000 for CorrLDA2, which is determined by monitoring in-sample log-likelihood convergence of topic and aspect assignments.

\section{Results}
We measure document classification accuracy in \autoref{subsec:results-lda}, to show that the topics learned by both LDA and CorrLDA2 are indicative of viewpoints. However, these topic-viewpoint associations are not modeled by LDA. As such, an end user obtains a list of topics without knowledge of which topics are associated with which viewpoints. We demonstrate how CorrLDA2 can solve this problem by leveraging the learned topic-aspect relations in \autoref{subsec:results-topic-aspect-relations}. Finally, we show that these topic-aspect relations consistently represent the correct topic-viewpoint associations in \autoref{subsec:results-consistency}.

\subsection{LDA}
\label{subsec:results-lda}

\begin{table*}[ht!]
\footnotesize
%\captionsetup{font=footnotesize}
\centering
\caption{Most probable words (lemmatized) of topics, inferred by LDA with $T=6$.}
\label{tab:lda-6}
\centerline{
\begin{tabular}{|c|c|l|}
\hline
Topic & \begin{tabular}[c]{@{}c@{}}SVM\\ weight\end{tabular} & \multicolumn{1}{c|}{Top words}                                                                                                                                                                         \\ \hline
% 4     & -4.96       & \begin{tabular}[c]{@{}l@{}}palestinian israeli israel international peace political occupation \\ people conflict one also public violence process end right force include \end{tabular} \\ \hline
% 0     & -1.56       & \begin{tabular}[c]{@{}l@{}}palestinian sharon state israeli minister election roadmap plan \\ government prime think political n't arafat want party see new \end{tabular}              \\ \hline
% 3     & 0.21        & \begin{tabular}[c]{@{}l@{}}israel settlement bank west gaza israeli settler fence wall \\ territory jerusalem water line security land strip build green area\end{tabular}                 \\ \hline
% 1     & 0.66        & \begin{tabular}[c]{@{}l@{}}state israel palestinian right jewish solution arab two israeli \\ refugee agreement return palestine issue one border 1967 peace land\end{tabular}                    \\ \hline
% 2     & 5.66        & \begin{tabular}[c]{@{}l@{}}israel american arafat arab bush iraq peace us war terrorism\\ israeli-palestinian leader year appear terrorist east even process \end{tabular}      \\ \hline
5     & -6.49       & \begin{tabular}[c]{@{}l@{}}palestinian israel israeli international state occupation right conflict people one law refugee \end{tabular} \\ \hline
1     & -1.81       & \begin{tabular}[c]{@{}l@{}}palestinian sharon israeli government election plan minister political roadmap security prime arafat \end{tabular}              \\ \hline  % arafat
4     & 0.20        & \begin{tabular}[c]{@{}l@{}}israeli people attack one war year world civilian religious medium suicide kill \end{tabular}                 \\ \hline
3     & 2.16        & \begin{tabular}[c]{@{}l@{}}american bush arafat israel arab us iraq conflict state war united administration \end{tabular}                    \\ \hline
6     & 2.58        & \begin{tabular}[c]{@{}l@{}}palestinian israeli peace israel process violence two public political leadership side term \end{tabular}      \\ \hline
2     & 3.35        & \begin{tabular}[c]{@{}l@{}}israel settlement state west bank jewish jerusalem arab gaza israeli fence line \end{tabular}      \\ \hline
\end{tabular}
}
\end{table*}

\begin{figure}[ht]
    \centering
    \scalebox{0.9}{\input{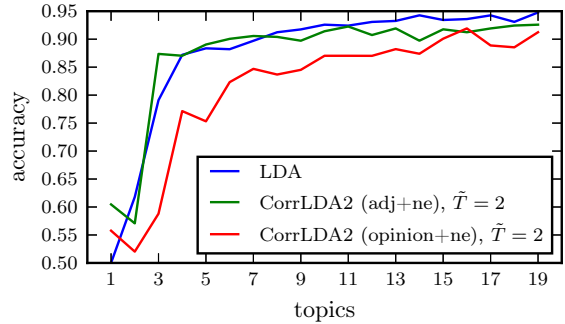}}
    \caption{Document classification accuracy using topics and aspects as features. The result shows that the topics learned by both LDA and CorrLDA2 are indeed associated with viewpoints, but LDA does not learn these topic-viewpoint associations. These associations are, however, learned completely unsupervised by the topic-aspect relations in CorrLDA2, and we show in \autoref{subsec:results-topic-aspect-relations} and \autoref{subsec:results-consistency} that these relations consistently represent correct topic-viewpoint associations.}
    \label{fig:lda-corrlda2-acc}
\end{figure}

\autoref{fig:lda-corrlda2-acc} shows that a document classification accuracy of $0.94$ can be achieved using 14 topics learned by LDA, indicating that topics are highly indicative of which viewpoint is being expressed. In other words, depending on which viewpoint one holds, one would then tend to bring up certain topics.

To demonstrate what the topics learned by LDA on this corpus look like, the most probable words of the topics inferred by LDA with $T=6$ are shown in \autoref{tab:lda-6}. Using these 6 topics as features, a document classification accuracy of $0.88$ can be achieved, as seen in \autoref{fig:lda-corrlda2-acc}. Hence, these topics are indicative of viewpoints, but LDA does not learn the topic-viewpoint associations. In a completely unsupervised setting without access to document-level viewpoint labels, and thus no knowledge of the feature weights learned by the SVM, it is then not clear how a human observer would manually associate each topic learned by LDA with a certain viewpoint, since LDA does not model these associations.

\subsection{CorrLDA2}
\label{subsec:results-corrlda2}

We observe that the most probable words of the topics learned by LDA in \autoref{tab:lda-6} usually include a nation, nationality or a political leader, such as: ``palestinian'', ``israel'', ``sharon'', ``arafat'' (Ariel Sharon was the Israeli prime minister and Yasser Arafat was the Palestinian president at the time when the documents in the corpus were written). In particular, the most probable word in each of the two Palestinian topics (Topics 5 and 1) is ``palestinian'', while two of the four topics that contribute to an Israeli viewpoint according to the SVM feature weights (Topics 4 and 2) have ``israeli'' and ``israel'' as the most probable word, indicating that people tend to talk about themselves. 

Based on this observation, we extend the opinion words used in previous studies, consisting of adjectives, verbs, and adverbs \cite{Fang2012,Thonet2016}, with named entities (locations, people, organizations, nationalities etc.), while nouns are used as topical words. We compare this (opinion+ne) partition of the vocabulary with other partitions in \autoref{subsec:results-consistency}, and quantitatively show that the (opinion+ne) partition indeed produces the most consistently correct topic-viewpoint associations.

For both the (opinion+ne) and (adj+ne) partitions of the vocabulary, we use $\tilde{T}=2$ aspects while varying the number of topics $T$ to measure the document classification accuracy. The results are shown in \autoref{fig:lda-corrlda2-acc}. We observe that the (adj+ne) partition has a classification accuracy comparable to LDA. The (opinion+ne) partition gives a slightly lower classification accuracy, since adjectives, adverbs, verbs and named entities are excluded from the topics, although a classification accuracy of 0.92 can still be achieved for this partition using $T=16$ topics.

Even higher classification accuracies can be obtained by increasing the number of aspects $\tilde{T}$. However, our goal is not to obtain the highest document classification accuracy in a supervised setting. This result is only used to show that the topics and aspects are indeed indicative of viewpoints. Our goal is to be able to form human interpretable representations of viewpoints in a completely unsupervised setting. In the following section, we demonstrate how this can be achieved by leveraging the model structure of CorrLDA2 to form groups of topics with an aspect. As a group, these topics and aspects can then be easier to interpret and assigned to a viewpoint by a human, than doing so for each topic individually. Since only 2 viewpoints are expressed in the Bitterlemons corpus, we set the number of aspects $\tilde{T}=2$, and show that each aspect is associated with a viewpoint.

% Topics are thus indicative of which viewpoint is being expressed, but we also see that in the absence the SVM feature weights obtained from labeled data, the process of assigning a topic to a viewpoint is not always clear for a human observer. In the following section, we demonstrate how the model structure of CorrLDA2 can be leveraged to form groups of topics with an aspect in a completely unsupervised setting. As a group, these topics and aspects can then be easier to interpret and assigned to a viewpoint by a human, than doing so for each topic individually.

\subsubsection{Topic-Aspect Relations}
\label{subsec:results-topic-aspect-relations}

\begin{table*}[t!]
\footnotesize
\captionsetup{font=footnotesize}
\centering
% Israeli viewpoint
\caption{Topics that co-occur with Aspect 1 with a frequency greater than 0.7. Learned by CorrLDA2 on the (opinion+ne) partition with $T=20$ topics and $\tilde{T}=2$ aspects.}
\label{tab:corrlda2-pal}
\centerline{
\begin{tabular}{l|c|l|}
\cline{2-3}
                                                       & \begin{tabular}[c]{@{}c@{}}SVM\\ weight\end{tabular} & \multicolumn{1}{c|}{Top words}\\ \hline
\rowcolor[HTML]{EFEFEF} 
\multicolumn{1}{|l|}{\cellcolor[HTML]{EFEFEF}Aspect 1} & -4.01     & palestinian israeli israel international political also jewish jerusalem palestine include authority arab                                              \\ \hline
\multicolumn{1}{|l|}{Topic 15}                         & -5.36     & \begin{tabular}[c]{@{}l@{}}occupation peace conflict side people government violence two process settlement confrontation land\end{tabular}            \\ \hline
\multicolumn{1}{|l|}{Topic 9}                          & -3.77     & \begin{tabular}[c]{@{}l@{}}peace process one role position leadership government change time situation effort community\end{tabular}                   \\ \hline
\multicolumn{1}{|l|}{Topic 11}                         & -1.70     & \begin{tabular}[c]{@{}l@{}}intifada people resistance struggle occupation movement time mean demonstration way use mass\end{tabular}                   \\ \hline
\multicolumn{1}{|l|}{Topic 14}                         & -1.58     & \begin{tabular}[c]{@{}l@{}}people land world one year peace day hope thing child suffering future\end{tabular}                                         \\ \hline
\multicolumn{1}{|l|}{Topic 16}                         & -0.97     & \begin{tabular}[c]{@{}l@{}}law right territory court wall issue violation decision case opinion convention occupation\end{tabular}                     \\ \hline
\multicolumn{1}{|l|}{Topic 5}                          & -0.47     & \begin{tabular}[c]{@{}l@{}}refugee right state return problem 242 resolution solution principle issue border 1967\end{tabular}                         \\ \hline
\multicolumn{1}{|l|}{Topic 3}                          & -0.28     & \begin{tabular}[c]{@{}l@{}}water economy issue aid resource security area infrastructure source time use worker\end{tabular}                           \\ \hline
\multicolumn{1}{|l|}{Topic 13}                         & 0.13      & \begin{tabular}[c]{@{}l@{}}state solution two one population territory citizen border 1967 people land majority\end{tabular}                           \\ \hline
\multicolumn{1}{|l|}{Topic 7}                          & 1.17      & \begin{tabular}[c]{@{}l@{}}settlement fence line wall separation border security land territory area withdrawal 1967\end{tabular}                      \\ \hline
\end{tabular}
}

\bigskip

% Palestinian viewpoint
\caption{Topics that co-occur with Aspect 2 with a frequency greater than 0.7. Learned by CorrLDA2 on the (opinion+ne) partition with $T=20$ topics and $\tilde{T}=2$ aspects.}
\label{tab:corrlda2-isr}
\centerline{
\begin{tabular}{l|c|l|}
\cline{2-3}
                                                       & \begin{tabular}[c]{@{}c@{}}SVM\\ weight\end{tabular} & \multicolumn{1}{c|}{Top words}                                                                                                                                                                     \\ \hline
\rowcolor[HTML]{EFEFEF} 
\multicolumn{1}{|l|}{\cellcolor[HTML]{EFEFEF}Aspect 2} & 4.01       & sharon israel arafat palestinian gaza american prime arab new us bush israeli                                                                         \\ \hline
\multicolumn{1}{|l|}{Topic 17}                         & 5.84       & \begin{tabular}[c]{@{}l@{}}peace process year two violence state security leader settlement way term terrorism\end{tabular}       \\ \hline
\multicolumn{1}{|l|}{Topic 20}                         & 2.72       & \begin{tabular}[c]{@{}l@{}}attack war terrorism suicide violence conflict civilian force bombing victory terror act\end{tabular}       \\ \hline
\multicolumn{1}{|l|}{Topic 19}                         & 1.71       & \begin{tabular}[c]{@{}l@{}}conflict administration president war region policy world leader minister pressure hand peace\end{tabular}       \\ \hline
\multicolumn{1}{|l|}{Topic 6}                          & 0.54       & \begin{tabular}[c]{@{}l@{}}election government party coalition public issue leadership majority position opposition campaign year\end{tabular}       \\ \hline
\multicolumn{1}{|l|}{Topic 10}                         & 0.39       & \begin{tabular}[c]{@{}l@{}}plan disengagement security withdrawal settlement part territory border state minister intention order\end{tabular}       \\ \hline
\multicolumn{1}{|l|}{Topic 8}                          & -0.32      & \begin{tabular}[c]{@{}l@{}}minister leader leadership one time intelligence security organization decision authority day faction\end{tabular}       \\ \hline
\end{tabular}
}
\end{table*}

In the sampling process of CorrLDA2, each opinion word is first assigned to a supertopic $x$, which is one of the $T$ topics that the topical words are assigned to. Then, one of the $\tilde{T}$ aspects is also assigned to the opinion word, conditioned on the supertopic. Since each opinion word is assigned to both a topic and an aspect, statistics of topic-aspect relations can be obtained by counting the number of times a topic co-occurs with an aspect. For the following experiment, we use the (opinion+ne) partition of the vocabulary, with $T = 20$ topics and $\tilde{T} = 2$ aspects. Using the condition that a topic is related to a certain aspect if they co-occur with a frequency greater than 0.7, the topics-aspect groups of \autoref{tab:corrlda2-pal} and \autoref{tab:corrlda2-isr} are obtained. Five out of the twenty topics do not pass the 0.7 threshold and can be viewed as neutral topics. %Even though the (adj+ne) partitioning of the vocabulary results in a higher classification accuracy of the documents, the (opinion+ne) partition yields more consistently correct topic-viewpoint associations, which is shown in \autoref{subsec:results-consistency}.

In these tables, the SVM feature weights for the aspects and the topics are obtained separately. The weights for the aspects are obtained by training a linear SVM with only the inferred aspect assignments of a document, while the weights for the topics are obtained using only the inferred topic assignments. The reason for using two separate processes is to avoid training the SVM with correlated features, which could degrade the significance of the learned feature weights.

We observe that in \autoref{tab:corrlda2-pal}, eight out of ten SVM weights are negative, indicating that a Palestinian viewpoint is learned, while all weights except one in \autoref{tab:corrlda2-isr} are positive, indicating an Israeli viewpoint. This result suggests that the topic-aspect relations learned by CorrLDA2 are strong indicators of the viewpoints that the topics are associated with.

Qualitatively, to provide a possible interpretation of the learned topics-aspect groups, we can see that the most probable word for Aspect 1 is ``palestinian'', while the two most probable words for Aspect 2 are ``sharon'' and ``israel''. This is in accordance with the observation made in \autoref{subsec:results-lda} that people tend to talk about themselves. Each aspect and topic individually might still be hard for a human to associate with either a Palestinian or an Israeli viewpoint. However, using the topic-aspect relations learned by CorrLDA2, we can now leverage all the topical and opinion words of a whole group of topics, together with an aspect, to associate the whole group with a viewpoint. For instance, certain keywords such as ``occupation'' is only used in the Palestinian group of topics, while ``terrorism'' is only used in the Israeli group.

In the Palestinian group of topics shown in \autoref{tab:corrlda2-pal}, the most probable word of Topic 15 include ``occupation'', ``violence'', ``settlement'', ``confrontation'', and ``land''. Topic 16 include the words ``territory'', ``violation'', and ``occupation''. These topics seem to refer to what Palestinians consider to be occupation of Palestinian territory, which they see as being in violation to international law. The top five words in Topic 11 include ``intifada'', ``resistance'', ``struggle'', and ``occupation''. Intifada is an Arabic word referring to resistance movements. During the period in which the documents of the corpus were published, what is generally called the Second Intifada, or the Al-Aqsa Intifada, took place. The top four words of Topic~5 include ``refugee'', ``right'', and ``return'', referring to what the Palestinians view as the Palestinian right of return to territory occupied by the Israelis, which has displaced millions of Palestinians.

In the Israeli group of topics shown in \autoref{tab:corrlda2-isr}, Topic 17 include the words ``violence'' and ``terrorism'' together with ``security'' and ``settlement''. The top four words of Topic 20 are: ``attack'', ``war'', ``terrorism'', and ``suicide'', referring to the numerous suicide attacks conducted by what the Israelis consider to be Palestinian terrorists during this period. Topic 10 contains the words ``disengagement'', ``withdrawal'', and ``settlement'', referring to the Israeli disengagement from Gaza, which the Israelis consider demonstrates their willingness to go through with the peace negotiations.

Again, this is just one possible way that an end user could interpret the results. However, this qualitative observation, combined with the quantitative results of the SVM weights for each aspect and topic, suggest that the topic-aspect relations learned by CorrLDA2 are indeed indicative of the viewpoint associations of the topics.

\subsubsection{Consistency of Topic-Aspect Relations}
\label{subsec:results-consistency}

\begin{figure*}[ht]
\centering
\begin{subfigure}[t]{.475\textwidth}
  \centering
  \scalebox{0.8}{\input{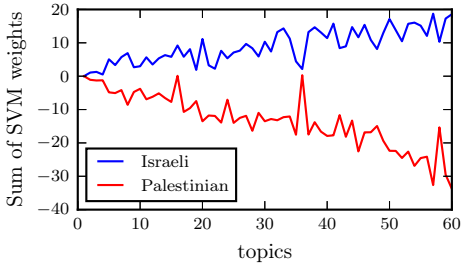}}
  \caption{(opinion+ne): adjectives, adverbs, verbs, and named entities.}
  \label{subfig:topicsums-pos-ner}
\end{subfigure}
\begin{subfigure}[t]{.475\textwidth}
  \centering
  \scalebox{0.8}{\input{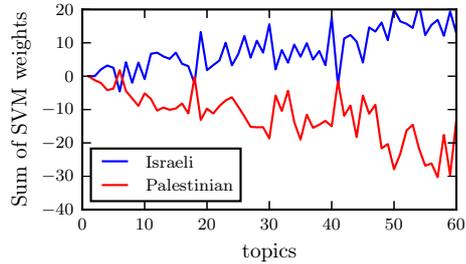}}
  \caption{(opinion): adjectives, adverbs, and verbs.}
  \label{subfig:topicsums-pos}
\end{subfigure}

\begin{subfigure}[b]{.475\textwidth}
  \centering
  \scalebox{0.8}{\input{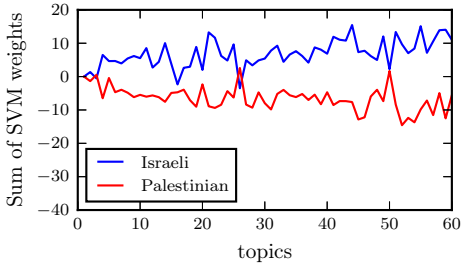}}
  \caption{(adj+ne): adjectives and named entities.}
  \label{subfig:topicsums-adj-ner}
\end{subfigure}
\begin{subfigure}[b]{.475\textwidth}
  \centering
  \scalebox{0.8}{\input{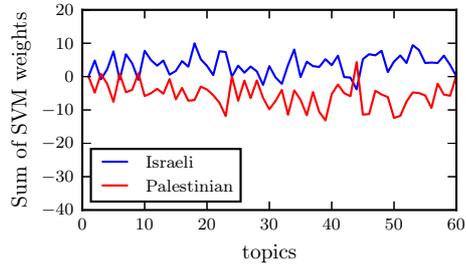}}
  \caption{(ne): only named entities.}
  \label{subfig:topicsums-ner}
\end{subfigure}
\caption{Sum of SVM weights of the topics in the topics-aspect groups, using different vocabulary partitions. The part-of-speech used as opinion words are listed below each figure.}
\label{fig:topicsums-supfig}
\end{figure*}

To evaluate how consistently the topic-aspect relations learned by CorrLDA2 can be used for associating topics to viewpoints, we form two topics-aspect groups according to the same procedure as described in \autoref{subsec:results-topic-aspect-relations}. For each of the two groups, if the aspect associated with the group has a negative SVM feature weight, then the group is classified as Palestinian. If the aspect has a positive weight, then it is classified as an Israeli group. We then define the score of a group to be the sum of the SVM feature weights of the topics in the group. As an example, we look at the case of using $T=20$ topics on the (opinion+ne) partition, which corresponds to the same situation as described in \autoref{subsec:results-topic-aspect-relations}. The resulting two topics-aspect groups are the ones shown in \autoref{tab:corrlda2-pal} and \autoref{tab:corrlda2-isr}. Since the aspect of \mbox{\autoref{tab:corrlda2-pal}} has a negative feature weight, this group is classified as Palestinian and its score is $-12.83$. The aspect of the group shown in \autoref{tab:corrlda2-isr} has a positive feature weight, and thus the group is classified as Israeli, with a score of $10.88$. We form two groups in this way for each $T=1,2, \ldots, 60$ total number of topics. We also compare the scores obtained from different partitions of the vocabulary, including the partition used in previous studies, with adjectives, adverbs, and verbs as opinion words \cite{Fang2012, Thonet2016}, which we refer to as (opinion), and a partition where the opinion words only consists of named entities, which we refer to as (ne). The resulting scores are shown in \autoref{fig:topicsums-supfig}.

We observe that the (opinion) partition, shown in \autoref{subfig:topicsums-pos}, produces a wider separation between the scores of the two topics-aspect groups compared to the (adj+ne) partition, shown in \autoref{subfig:topicsums-adj-ner}. However, for some number of topics in both of these partitions, the Palestinian line crosses the Israeli line, meaning that many of the learned topic-aspect relations do not correspond to the correct topic-viewpoint associations. Using only named entities as opinion words results in the worst performance, with multiple overlaps and small separation between the scores of the two groups, as seen in \autoref{subfig:topicsums-ner}. Using a combination of the opinion words proposed by previous studies (adjectives, adverbs, and verbs) together with named entities, the scores shown in \autoref{subfig:topicsums-pos-ner} are obtained. This vocabulary partition produces the most consistently correct topic-viewpoint associations, with a clear separation and no overlap between the obtained Israeli and Palestinian scores.

\section{Conclusion}
We introduce a novel approach of applying CorrLDA2 to the task of viewpoint and topic modeling. The vocabulary is partitioned to obtain bimodal data, where one modality consists of topical words (nouns), and the other modality consists of opinion words (adjectives, adverbs, verbs, and named entities). We show that by including named entities to the set of opinion words, the obtained topic-viewpoint associations become more consistently correct compared to the opinion words used in previous studies. Using the learned relationships between the modalities, groups of topics-aspect are formed, creating human interpretable representations of viewpoints. We also introduce a principled method to evaluate which viewpoint a topic or aspect is associated with, by leveraging the magnitudes and signs of the feature weights of a linear SVM. With this, we show, both quantitatively and qualitatively, that the learned groups of topics-aspect are contextually coherent, and consistently correctly associated with the viewpoints.

%One of the strengths of CorrLDA2 compared to previously proposed approaches for viewpoint modeling is that CorrLDA2 does not constrain a topic to be related to a constant number of specific aspects. Instead, any topic can be related to any number of aspects. Also, there is no constraint to the number of aspects $\tilde{T}$. As such, we hypothesize that the model can be applied to any multi-subject corpus, i.e. a collection of documents that is not restricted to a single subject such as the Israeli-Palestinian conflict. This includes any collection of news articles, and it would be interesting to see the topic-aspect relations that are learned on such a corpus in future studies.

\section{Future Work}
Unlike previously proposed models, CorrLDA2 does not assume the viewpoints to be known a priori, relationships between topics and aspects are learned, aspects can be associated with multiple topics, and it does not require the number of topics and aspects to be equal. As such, we hypothesize that the model can be applied to any multi-subject corpus, i.e. a collection of documents that is not restricted to a single subject such as the Israeli-Palestinian conflict. This includes any collection of news articles.

If more than two viewpoints are expressed in the corpus, then the quantitative evaluation method introduced in this paper has to be extended. Standard techniques for multiclass classification could potentially be applied, such as one-vs-all or one-vs-one classification strategy.

% There are several ways that the quantitative evaluation method, which is based on interpreting the feature weights of a linear SVM, can be extended. One possibility is to use a one-vs-all classification strategy by separately training $L$ classifiers, where $L$ is the number of viewpoints expressed in the corpus. If we let viewpoint $l$ have a label of $+1$ in the $l$th classifier, then the features that are associated with this viewpoint are the ones with a positive feature weight in this particular classifier. In a one-vs-one approach, the average feature weights obtained from the classifiers involving viewpoint $l$ could be leveraged.

Another domain that the results of this study can be applied to is information retrieval, where the topics-aspect groups learned by CorrLDA2 can be used to query for documents of a specific viewpoint.

Improvements to the model could potentially be achieved by introducing sentence-level topics, as done in other related studies \cite{Jo2011,Thonet2016}. The supertopics of the opinion words would then be the sentence-level topic, which would enforce an even greater relationship between topics and aspects.

The part-of-speech categories are used as proxies for modeling topical and opinion words. By moving the category of named entities from the set of topical words to the set of opinion words, we show that the topic-viewpoint associations become more consistently correct. This invites further study of how better vocabulary partitions can be obtained. Nouns may well indicate an opinion; verbs, adverbs, and adjectives may well be topical rather than aspectual. An alternative to part-of-speech categories could be to leverage the vector spaces learned in word embeddings.

\section{Acknowledgments}
We would like to thank Thibaut Thonet for being part of some very interesting discussions regarding viewpoint modeling using statistical topic models.

\bibliographystyle{plain}
\bibliography{99_kexbib}{}

\end{document}